\documentclass[11pt,a4paper]{article}
\usepackage[hyperref]{acl2018}
\usepackage{times}
\usepackage{latexsym}

\usepackage{url}
\usepackage{multirow}
\usepackage{epsfig}
\usepackage{amsmath}
\usepackage{subcaption}
\usepackage{graphicx}
\usepackage{bm}
\usepackage{CJK}
\newcommand*{\affaddr}[1]{#1} 
\newcommand*{\affmark}[1][*]{\textsuperscript{#1}}
\newcommand*{\email}[1]{\texttt{#1}}

\aclfinalcopy 
\def\aclpaperid{1546} 

\title{Autoencoder as Assistant Supervisor: Improving Text Representation for Chinese Social Media Text Summarization}

\author{Shuming Ma\affmark[1], Xu Sun\affmark[1,2], Junyang Lin\affmark[3], Houfeng Wang\affmark[1]\\
\affaddr{\affmark[1]MOE Key Lab of Computational Linguistics, School of EECS, Peking University}\\
\affaddr{\affmark[2]Deep Learning Lab, Beijing Institute of Big Data Research, Peking University}\\
\affaddr{\affmark[3]School of Foreign Languages, Peking University}\\
\email{\{shumingma, xusun, linjunyang, wanghf\}@pku.edu.cn}\\
}

\date{}

\begin{document}
\maketitle
\begin{abstract}
Most of the current abstractive text summarization models are based on the sequence-to-sequence model (Seq2Seq). The source content of social media is long and noisy, so it is difficult for Seq2Seq to learn an accurate semantic representation. Compared with the source content, the annotated summary is short and well written. Moreover, it shares the same meaning as the source content. In this work, we supervise the learning of the representation of the source content with that of the summary. In implementation, we regard a summary autoencoder as an assistant supervisor of Seq2Seq. Following previous work, we evaluate our model on a popular Chinese social media dataset. Experimental results show that our model achieves the state-of-the-art performances on the benchmark dataset.\footnote{The code is available at \url{https://github.com/lancopku/superAE}}
\end{abstract}

\section{Introduction}

Text summarization is to produce a brief summary of the main ideas of the text. Unlike extractive text summarization~\citep{extra04,extra10,discourse}, which selects words or word phrases from the source texts as the summary, abstractive text summarization learns a semantic representation to generate more human-like summaries. Recently, most models for abstractive text summarization are based on the sequence-to-sequence model, which encodes the source texts into the semantic representation with an encoder, and generates the summaries from the representation with a decoder. 

The contents on the social media are long, and contain many errors, which come from spelling mistakes, informal expressions, and grammatical mistakes~\citep{noisy}. Large amount of errors in the contents cause great difficulties for text summarization. As for RNN-based Seq2Seq, it is difficult to compress a long sequence into an accurate representation~\citep{hierarchical}, because of the gradient vanishing and exploding problem.

Compared with the source content, it is easier to encode the representations of the summaries, which are short and manually selected. Since the source content and the summary share the same points, it is possible to supervise the learning of the semantic representation of the source content with that of the summary. 

In this paper, we regard a summary autoencoder as an assistant supervisor of Seq2Seq. 
First, we train an autoencoder, which inputs and reconstructs the summaries, to obtain a better representation to generate the summaries. Then, we supervise the internal representation of Seq2Seq with that of autoencoder by minimizing the distance between two representations. Finally, we use adversarial learning to enhance the supervision.
Following the previous work~\citep{MaEA2017}, We evaluate our proposed model on a Chinese social media dataset. Experimental results show that our model outperforms the state-of-the-art baseline models. More specifically, our model outperforms the Seq2Seq baseline by the score of 7.1 ROUGE-1, 6.1 ROUGE-2, and 7.0 ROUGE-L.

\section{Proposed Model}

We introduce our proposed model in detail in this section.

\subsection{Notation}\label{overview}

Given a summarization dataset that consists of $N$ data samples, the $i^{th}$ data sample ($x_{i}$, $y_{i}$) contains a source content $x_{i}=\{x_{1},x_{2},...,x_{M}\}$, and a summary $y_{i}=\{y_{1},y_{2},...,y_{L}\}$, while $M$ is the number of the source words, and $L$ is the number of the summary words. At the training stage, we train the model to generate the summary $\bm{y}$ given the source content $\bm{x}$. At the test stage, the model decodes the predicted summary $\bm{y'}$ given the source content $\bm{x}$.

\begin{figure}[tb]
	\centering
	\subcaptionbox{Training Stage}{\includegraphics[width=1.0\linewidth]{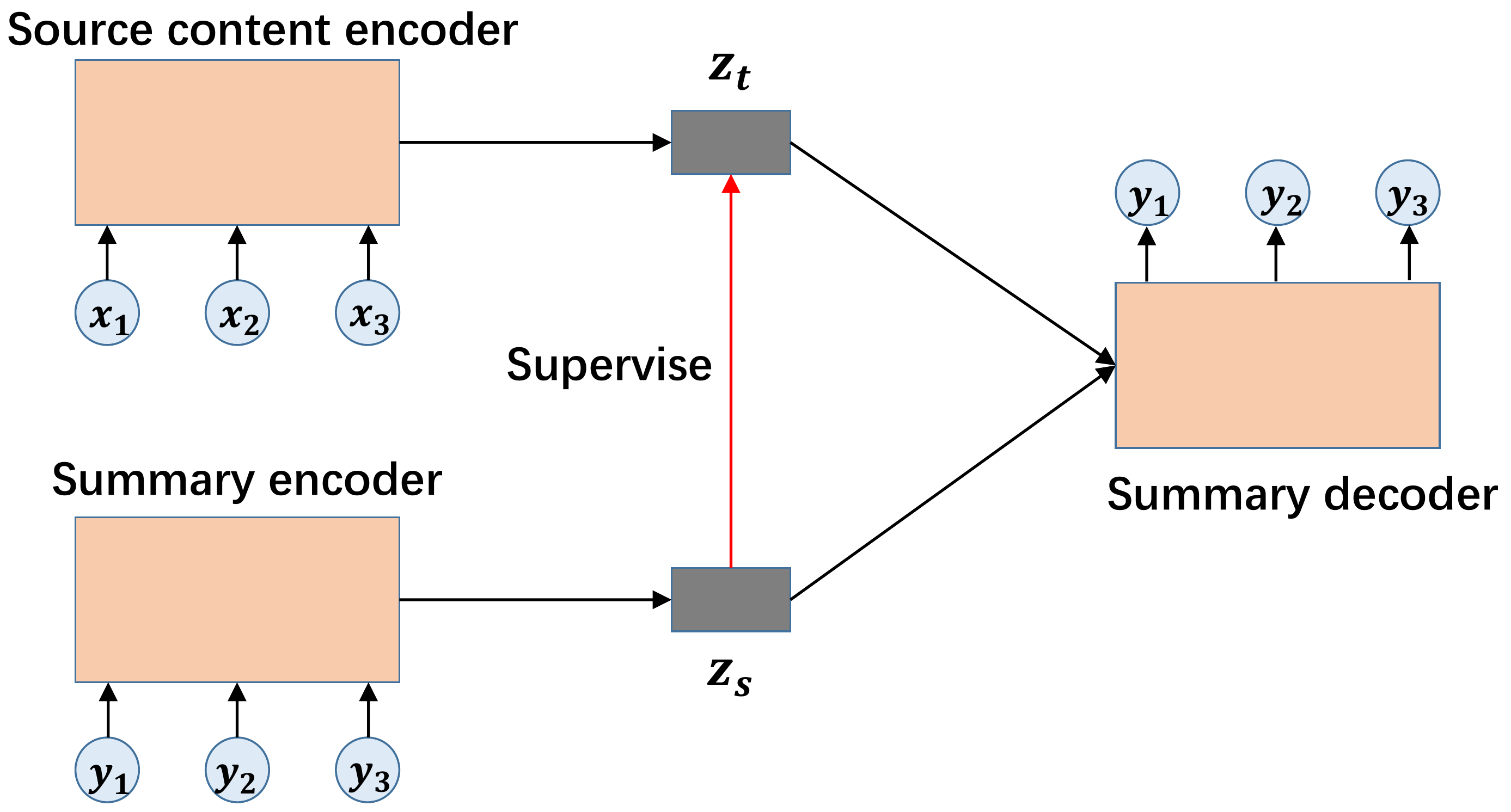}} \\
	\subcaptionbox{Test Stage}{\includegraphics[width=1.0\linewidth]{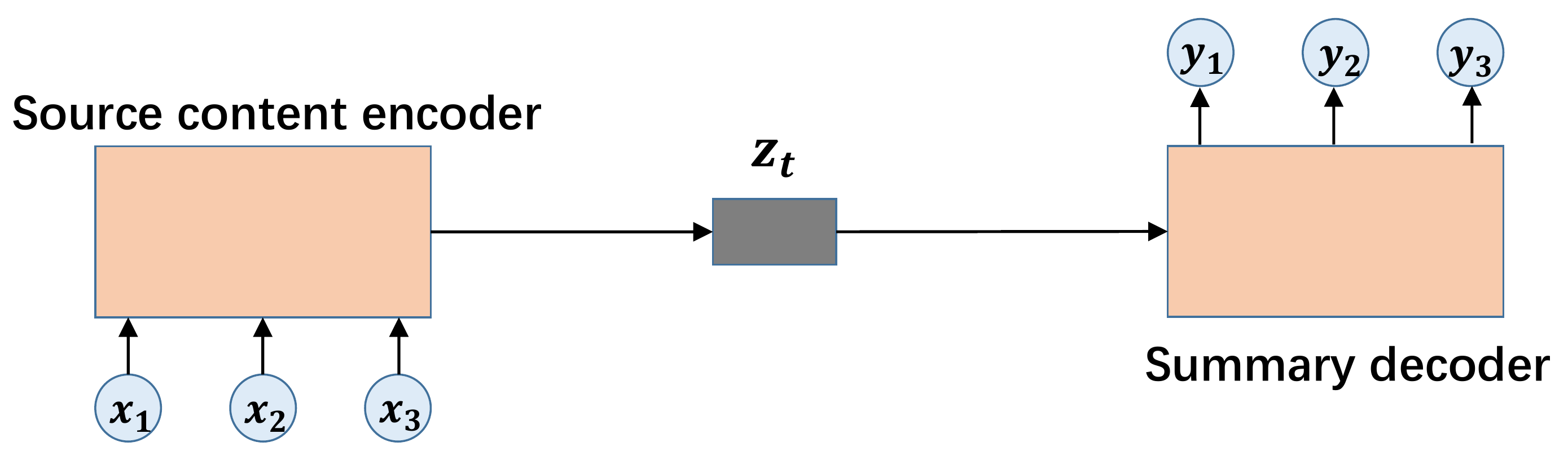}}
	\caption{The overview of our model. The model consists of a sequence-to-sequence model and an autoencoder model. At the training stage, we use the autoencoder to supervise the sequence-to-sequence model. At the test stage, we use the sequence-to-sequence model to generate the summaries.}\label{model_fig}
	\vspace{-0.05in}
\end{figure}

\subsection{Supervision with Autoencoder}\label{supervise}

Figure~\ref{model_fig} shows the architecture of our model. At the training stage, the source content encoder compresses the input contents $\bm{x}$ into the internal representation $z_t$ with a Bi-LSTM encoder. At the same time, the summary encoder compresses the reference summary $\bm{y}$ into the representation $z_s$. Then both $z_t$ and $z_s$ are fed into a LSTM decoder to generate the summary. Finally, the semantic representation of the source content is supervised by the summary.

We implement the supervision by minimizing the distance between the semantic representations $z_t$ and $z_s$, and this term in the loss function can be written as:
\begin{equation}
L_S = \frac{\lambda}{N_h} d(z_t,z_s)\label{loss_supervision}
\end{equation}
where $d(z_t,z_s)$ is a function which measures the distance between $z_s$ and $z_t$. $\lambda$ is a tunable hyper-parameter to balance the loss of the supervision and the other parts of the loss, and $N_h$ is the number of the hidden unit to limit the magnitude of the distance function. We set $\lambda=0.3$ based on the performance on the validation set. The distance between two representations can be written as:
\begin{equation}
d(z_t,z_s)=\lVert z_t-z_s \rVert_2
\end{equation}

\subsection{Adversarial Learning}\label{adversarial}

We further enhance the supervision with the adversarial learning approach. As shown in Eq.~\ref{loss_supervision}, we use a fixed hyper-parameter as a weight to measure the strength of the supervision of the autoencoder. However, in the case when the source content and summary have high relevance, the strength of the supervision should be higher, and when the source content and summary has low relevance, the strength should be lower. In order to determine the strength of supervision more dynamically, we introduce the adversarial learning. 
More specifically, we regard the representation of the autoencoder as the ``gold'' representation, and that of the sequence-to-sequence as the ``fake'' representation. A model is trained to discriminate between the gold and fake representations, which is called a discriminator. The discriminator tries to identify the two representations. On the contrary, the supervision, which minimizes the distance of the representations and makes them similar, tries to prevent the discriminator from making correct predictions. In this way, when the discriminator can distinguish the two representations (which means the source content and the summary has low relevance), the strength of supervision will be decreased, and when the discriminator fails to distinguish, the strength of supervision will be improved. 

In implementation of the adversarial learning, the discriminator objective function can be written as:
\begin{equation}\label{discriminator}
\begin{split}
L_D(\theta_D)=
& -\log{P_{\theta_D}(y=1|z_t)} \\
& -\log{P_{\theta_D}(y=0|z_s)}
\end{split}
\end{equation}
where $P_{\theta_D}(y=1|z)$ is the probability that the discriminator identifies the vector $z$ as the ``gold'' representation, while $P_{\theta_D}(y=0|z)$ is the probability that the vector $z$ is identified as the ``fake'' representation, and $\theta_D$ is the parameters of the discriminator. When minimizing the discriminator objective, we only train the parameters of the discriminator, while the rest of the parameters remains unchanged.

The supervision objective to be against the discriminator can be written as:
\begin{equation}\label{supervisor}
\begin{split}
L_G(\theta_E)=
& -\log{P_{\theta_D}(y=0|z_t)} \\
& -\log{P_{\theta_D}(y=1|z_s)}
\end{split}
\end{equation}
When minimizing the supervision objective, we only update the parameters of the encoders.

\subsection{Loss Function and Training}\label{training}

There are several parts of the objective functions to optimize in our models. The first part is the cross entropy losses of the sequence-to-sequence and the autoencoder:
\begin{equation}\label{cross_seq2seq}
L_{Seq2seq} = -\sum_{i=1}^{N}p_{Seq2seq}(y_i|z_s)
\end{equation}
\begin{equation}\label{cross_ae}
L_{AE} = -\sum_{i=1}^{N}p_{AE}(y_i|z_t)
\end{equation}
The second part is the L2 loss of the supervision, as written in Equation~\ref{loss_supervision}. The last part is the adversarial learning, which are Equation~\ref{discriminator} and Equation~\ref{supervisor}. The sum of all these  parts is the final loss function to optimize.

We use the Adam~\cite{KingmaBa2014} optimization method to train the model. For the hyper-parameters of Adam optimizer, we set the learning rate $\alpha = 0.001$, two momentum parameters $\beta_{1} = 0.9$ and $\beta_{2} = 0.999$
respectively, and $\epsilon = 1 \times 10^{-8}$. We clip the gradients~\cite{gradientclip} to the maximum norm of 10.0.


\section{Experiments}

Following the previous work~\citep{MaEA2017}, we evaluate our model on a popular Chinese social media dataset. We first introduce the datasets, evaluation metrics, and experimental details. Then, we compare our model with several state-of-the-art systems.

\subsection{Dataset}

\noindent\textbf{Large Scale Chinese Social Media Text Summarization Dataset (LCSTS)} is constructed by \citet{lcsts}. The dataset consists of more than 2,400,000 text-summary pairs, constructed from a famous Chinese social media website called Sina Weibo.\footnote{\url{http://weibo.com}} It is split into three parts, with 2,400,591 pairs in PART I, 10,666 pairs in PART II and 1,106 pairs in PART III. All the text-summary pairs in PART II and PART III are manually annotated with relevant scores ranged from 1 to 5. We only reserve pairs with scores no less than 3, leaving 8,685 pairs in PART II and 725 pairs in PART III. Following the previous work~\cite{lcsts}, we use PART I as training set, PART II as validation set, and PART III as test set. 

\subsection{Evaluation Metric}

Our evaluation metric is ROUGE score~\cite{rouge}, which is popular for summarization evaluation. The metrics compare an automatically produced summary with the reference summaries, by computing overlapping lexical units, including unigram, bigram, trigram, and longest common subsequence (LCS). Following previous work~\cite{abs,lcsts}, we use ROUGE-1 (unigram), ROUGE-2 (bi-gram) and ROUGE-L (LCS) as the evaluation metrics in the reported experimental results.

\subsection{Experimental Details}

The vocabularies are extracted from the training sets, and the source contents and the summaries share the same vocabularies. In order to alleviate the risk of word segmentation mistakes, we split the Chinese sentences into characters. We prune the vocabulary size to 4,000, which covers most of the common characters. 

We tune the hyper-parameters based on the ROUGE scores on the validation sets. 
We set the word embedding size and the hidden size to 512, and the number of LSTM layers is 2. The batch size is 64, and we do not use dropout~\cite{dropout} on this dataset. Following the previous work~\cite{DRGD}, we implement the beam search, and set the beam size to 10.

\subsection{Baselines}

We compare our model with the following state-of-the-art baselines.

\begin{itemize}
\item \textbf{RNN} and \textbf{RNN-cont} are two sequence-to-sequence baseline with GRU encoder and decoder, provided by~\citet{lcsts}. The difference between them is that RNN-context has attention mechanism while RNN does not.
\item \textbf{RNN-dist}~\cite{distraction} is a distraction-based neural model, which the attention mechanism focuses on the different parts of the source content.
\item \textbf{CopyNet}~\cite{copynet} incorporates a copy mechanism to allow parts of the generated summary are copied from the source content.
\item \textbf{SRB}~\cite{MaEA2017} is a sequence-to-sequence based neural model with improving the semantic relevance between the input text and the output summary.
\item \textbf{DRGD}~\cite{DRGD} is a deep recurrent generative decoder model, combining the decoder with a variational autoencoder.
\item \textbf{Seq2seq} is our implementation of the sequence-to-sequence model with the attention mechanism, which  has the same experimental setting as our model for fair comparison.
\end{itemize}

\begin{table}[t]
	\centering
	\begin{tabular}{@{}l@{}c@{}c@{}c@{}}
		\hline
		\multicolumn{1}{l}{\textbf{Models}} &
		\multicolumn{1}{c}{\textbf{R-1}} & 
		\multicolumn{1}{c}{\textbf{R-2}} &  
		\multicolumn{1}{c}{\textbf{R-L}}   \\ \hline
		RNN-W\cite{lcsts} &  17.7  & 8.5  &  15.8 \\
		RNN\cite{lcsts} & 21.5 & 8.9 & 18.6 \\
		RNN-cont-W\cite{lcsts} & 26.8  & 16.1  &  24.1  \\
		RNN-cont\cite{lcsts} & 29.9 & 17.4 & 27.2 \\
		SRB\cite{MaEA2017} & 33.3 & 20.0 & 30.1 \\
		CopyNet-W\cite{copynet} & 35.0 & 22.3 & 32.0 \\ 
		CopyNet\cite{copynet} & 34.4 & 21.6  &  31.3 \\
		RNN-dist\cite{distraction} & 35.2 & 22.6 & 32.5 \\
		DRGD\cite{DRGD} & 37.0 & 24.2 & 34.2  \\ \hline
		Seq2Seq (our impl.) & 32.1 & 19.9 & 29.2 \\
		\textbf{+superAE (this paper)} & \textbf{39.2} & \textbf{26.0} & \textbf{36.2} \\
        w/o adversarial learning & 37.7 & 25.3 & 35.2 \\ \hline
		
	\end{tabular}
	\caption{Comparison with state-of-the-art models on the LCSTS test set. R-1, R-2, and R-L denote ROUGE-1, ROUGE-2, and ROUGE-L, respectively. The models with a suffix of `W' in the table are word-based, while the rest of models are character-based.}\label{tab_lcsts_sota}
\end{table}

\subsection{Results}

For the purpose of simplicity, we denote our supervision with autoencoder model as \textbf{superAE}. We report the ROUGE F1 score of our model and the baseline models on the test sets. 

Table~\ref{tab_lcsts_sota} summarizes the results of our superAE model and several baselines.
We first compare our model with Seq2Seq baseline. It shows that our superAE model has a large improvement over the Seq2Seq baseline by 7.1 ROUGE-1, 6.1 ROUGE-2, and 7.0 ROUGE-L, which demonstrates the efficiency of our model. Moreover, we compare our model with the recent summarization systems, which have been evaluated on the same training set and the test sets as ours. Their results are directly reported in the referred articles. It shows that our superAE outperforms all of these models, with a relative gain of 2.2 ROUGE-1, 1.8 ROUGE-2, and 2.0 ROUGE-L. We also perform ablation study by removing the adversarial learning component, in order to show its contribution. It shows that the adversarial learning improves the performance of 1.5 ROUGE-1, 0.7 ROUGE-2, and 1.0 ROUGE-L.

We also give a summarization examples of our model. As shown in Table~\ref{model_example}, the SeqSeq model captures the wrong meaning of the source  content, and produces the summary that ``\emph{China United Airlines exploded in the airport}''. Our superAE model captures the correct points, so that the generated summary is close in meaning to the reference summary.

\begin{table}[t]
	\centering
	\normalsize
	\begin{tabular}{p{2.8cm}ll}
		\hline
		\multicolumn{1}{l}{\textbf{Models}} &
		\multicolumn{1}{c}{\textbf{2-class (\%)}} & 
		\multicolumn{1}{c}{\textbf{5-class (\%)}}   \\ \hline
		Seq2seq  & 80.7 &  65.1 \\
		\textbf{+superAE}  & \textbf{88.8 (+8.1)} & \textbf{71.7 (+6.6)}  \\ \hline
		
	\end{tabular}
	\caption{Accuracy of the sentiment classification on the Amazon dataset. We train a classifier which inputs internal representation provided by the sequence-to-sequence model, and outputs a predicted label. We compute the 2-class and 5-class accuracy of the predicted labels to evaluate the quality of the text representation. }\label{tab_amazon_acc}
\end{table}

\subsection{Analysis of text representation}

We want to analyze whether the internal text representation is improved by our superAE model. Since the text representation is abstractive and hard to evaluate, we translate the representation into a sentiment score with a sentiment classifier, and evaluate the quality of the representation by means of the sentiment accuracy.

We perform experiments on the Amazon Fine Foods Reviews Corpus~\cite{McAuley2013}. The Amazon dataset contains users' rating labels as well as the summary for the reviews, making it possible to train a classifier to predict the sentiment labels and a seq2seq model to generate summaries. First, we train the superAE model and the seq2seq model with the text-summary pairs until convergence. Then, we transfer the encoders to a sentiment classifier, and train the classifier with fixing the parameters of the encoders. The classifier is a simple feedforward neural network which maps the representation into the label distribution. Finally, we compute the accuracy of the predicted 2-class labels and 5-class labels. 

As shown in Table~\ref{tab_amazon_acc}, the seq2seq model achieves 80.7\% and 65.1\% accuracy of 2-class and 5-class, respectively. Our superAE model outperforms the baselines with a large margin of 8.1\% and 6.6\%. 

\begin{CJK}{UTF8}{gbsn}
\begin{table}[t]
\centering
    \begin{tabular}{p{7.2cm}}
    \hline
    \textbf{Source:}  昨晚，中联航空成都飞北京一架航班被发现有多人吸烟。后因天气原因，飞机备降太原机场。有乘客要求重新安检，机长决定继续飞行，引起机组人员与未吸烟乘客冲突。\\
Last night, several people were caught to smoke on a flight of China United Airlines from Chendu to Beijing. Later the flight temporarily landed on Taiyuan Airport. Some passengers asked for a security check but were denied by the captain, which led to a collision between crew and passengers.\\
    \hline
    \textbf{Reference:} 航班多人吸烟机组人员与乘客冲突。 \\
    Several people smoked on a flight which led to a collision between crew and passengers. \\
    \hline
    \textbf{Seq2Seq:}  中联航空机场发生爆炸致多人死亡。\\
	China United Airlines exploded in the airport, leaving several people dead.  \\
    \hline
    \textbf{+superAE:} 成都飞北京航班多人吸烟机组人员与乘客冲突。 \\
	Several people smoked on a flight from Chendu to Beijing, which led to a collision between crew and passengers.
\\
    \hline
    \end{tabular}
    \caption{A summarization example of our model, compared with Seq2Seq and the reference.}
    \label{model_example}
    \vspace{-3mm}
\end{table}
\end{CJK}

\section{Related Work}

\citet{abs} first propose an abstractive based summarization model, which uses an attentive CNN encoder to compress texts and a neural network language model to generate summaries.
\citet{ras} explore a recurrent structure for abstractive summarization. 
To deal with out-of-vocabulary problem, \citet{ibmsummarization} propose a generator-pointer model so that the decoder is able to generate words in source texts.
\citet{copynet} also solved this issue by incorporating copying mechanism, allowing parts of the summaries are copied from the source contents.
\citet{SeeEA2017} further discuss this problem, and incorporate the pointer-generator model with the coverage mechanism.
\citet{lcsts} build a large corpus of Chinese social media short text summarization, which is one of our benchmark datasets.
\citet{distraction} introduce a distraction based neural model, which forces the attention mechanism to focus on the difference parts of the source inputs.
\citet{MaEA2017} propose a neural model to improve the semantic relevance between the source contents and the summaries.

Our work is also related to the sequence-to-sequence model~\cite{ChoEA2014}, and the autoencoder model~\cite{bengio,ae1,ae2}. 
Sequence-to-sequence model is one of the most successful generative neural model, and is widely applied in machine translation~\cite{seq2seq,mapattention,stanfordattention}, text summarization~\cite{abs,ras,ibmsummarization}, and other natural language processing tasks.
Autoencoder~\cite{bengio} is an artificial neural network used for unsupervised learning of efficient representation.
Neural attention model is first proposed by~\citet{attention}. 

\section{Conclusion}

We propose a novel model, in which the autoencoder is a supervisor of the sequence-to-sequence model, to learn a better internal representation for abstractive summarization. An adversarial learning approach is introduced to further improve the supervision of the autoencoder. Experimental results show that our model outperforms the sequence-to-sequence baseline by a large margin, and achieves the state-of-the-art performances on a Chinese social media dataset.

\section*{Acknowledgements}

Our work is supported by National Natural Science Foundation of China (No. 61433015, No. 61673028), National High Technology Research and Development Program of China (863 Program, No. 2015AA015404), and the National Thousand Young Talents Program. Xu Sun is the corresponding author of this paper.

\nocite{SunEA2017,SunWei2017,dnerre,amr,discourse,wean,unpair}

\bibliographystyle{acl_natbib}
\bibliography{aeseq}

\end{document}